\documentclass{article}

\pdfoutput=1
\usepackage{arxiv}
\usepackage[toc,page]{appendix}
\usepackage[utf8]{inputenc} 
\usepackage[T1]{fontenc}    
\usepackage{hyperref}       
\usepackage{url}            
\usepackage{bm}
\usepackage{booktabs, makecell}       
\usepackage{amsfonts}       
\usepackage{nicefrac}       
\usepackage{microtype}      
\usepackage{lipsum}
\usepackage{biblatex}
\usepackage{comment}
\usepackage{graphicx}
\usepackage{amsmath}
\usepackage{caption}
\usepackage{subcaption}
\usepackage{multirow}
\usepackage{array}
\newcolumntype{C}[1]{>{\centering\arraybackslash}p{#1}}

\title{Measuring the False Sense of Security}

\author{
  Carlos Gomes\\
  ETH Zurich \\
  cagomes@ethz.ch \\
}

\begin{document}
\maketitle

\begin{abstract} 
Recently, several papers have demonstrated how widespread gradient masking is amongst proposed adversarial defenses. Defenses that rely on this phenomenon are considered failed, and can easily be broken. Despite this, there has been little investigation into ways of measuring the phenomenon of gradient masking and enabling comparisons of its extent amongst different networks. In this work, we investigate gradient masking under the lens of its mensurability, departing from the idea that it is a binary phenomenon. We propose and motivate several metrics for it, performing extensive empirical tests on defenses suspected of exhibiting different degrees of gradient masking. These are computationally cheaper than strong attacks, enable comparisons between models, and do not require the large time investment of tailor-made attacks for specific models. Our results reveal metrics that are successful in measuring the extent of gradient masking across different networks.
\end{abstract}

\section{Introduction}
Adversarial robustness has emerged as a prominent field in deep learning, studying the phenomenon of adversarial examples - small perturbations to the inputs of Deep Neural Networks that cause misclassifications to completely different classes. Numerous attacks have been put forward to find such perturbations in networks, as well as defenses to make networks robust to them. Within adversarial robustness, the topic of gradient masking has recently arisen.

Gradient masking is the phenomenon whereby a network's gradients are not useful for finding adversarial examples (e.g. Table \ref{grad_mask_toy}). Without informative gradients, optimization-based attacks are less effective. Some proposed defenses actually produce models that mask their gradients. Defenses that inherently rely on gradient masking are problematic as they target weaknesses in some attacks, rather than eliminating adversarial examples. These adversarial examples are often not very difficult to find through adapting the original attack to the defense at hand \cite{papernot2016science, tramerAdaptive, obfuscated}. In these cases, whether intentionally or not, the design of these defenses caused them to resort to gradient masking. However, there are also cases where the network appears to learn to mask its gradients without a design feature that can be obviously blamed. In these cases, the model learns to make some attacks weaker. This phenomenon has been likened to reward hacking from reinforcement learning \cite{ensemble}, and raises the troubling issue that a defense may rely on gradient masking despite the designers’ best efforts to avoid it.

\begin{table}[ht]
\centering
\begin{tabular}{ccc}
  \includegraphics[width=0.15\linewidth]{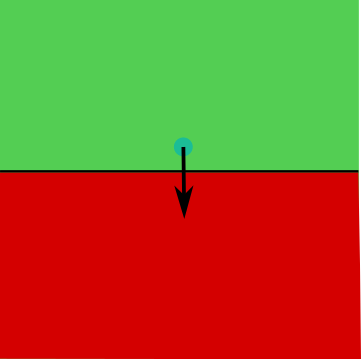} &   \includegraphics[width=0.15\linewidth]{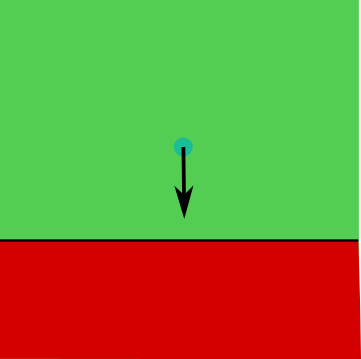} &
  \includegraphics[width=0.15\linewidth]{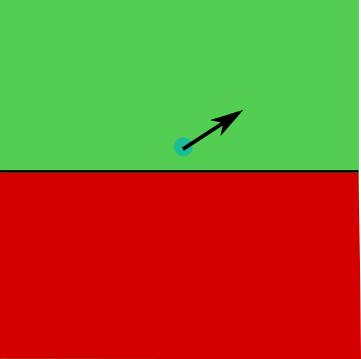}
  \\
(a) Undefended & (b) Robust & (c) Gradient Masking \\[6pt]
\end{tabular}
\caption{Diagram visualizing gradient masking. The blue circle represents a data point, and the arrow the direction of the gradient. The green and red areas represent different classification regions, separated by the decision boundary. (a) A model which is not robust. (b) For a truly robust model, we require that the distance from the data point to the boundary be increased. (c) A gradient masked model, which instead changes the direction of the gradient. The adversarial perturbation vector still exists.}
\label{grad_mask_toy}
\end{table}

Despite the undesirable properties of gradient masking, there has been little investigation on ways to detect and measure it. Currently, when we suspect a network of masking its gradients, we proceed with the often non-obvious and time consuming task of designing an attack that is strong enough to find adversarial examples that standard gradient-based attacks fail to find. Indeed, one can be left in the unfortunate situation where one is not sure if a network is masking its gradients, since the right attack may simply have not been found yet. We envision a metric for gradient masking, akin to the validation accuracy for overfitting, that could facilitate the evaluation of gradient masking in a network. Analogously to overfitting, such a metric would allow us to analyze gradient masking more quantitatively, beyond its view as a binary phenomenon. Gradient masking could be detected without the need for individually crafted attacks, and a better comparison of the extent of gradient masking across different networks would be possible. 

Our contributions are the following:
\begin{itemize}
    \item We depart from the idea of gradient masking as a binary feature.
    \item We propose, motivate and investigate several metrics to quantify gradient masking in neural networks. 
    \item We test these metrics on defenses which exhibit different degrees of this problem.
    \item We demonstrate that these metrics offer an insightful and computationally cheap solution to analyzing gradient masking. 
\end{itemize}

We now briefly highlight some related work. Since the initial definition of gradient masking by \cite{papernotPractical}, many works have explored the topic. In \cite{obfuscated}, the authors focus on cases of defenses proposed by the research community whose design masks gradients, so called "obfuscated gradients", and effectively break them. In \cite{ensemble} it is proposed that "adversarial training with single-step methods admits a degenerate global minimum, wherein the model’s loss can not be reliably approximated by a linear function", offering some explanation for the phenomenon of gradient masking. The authors also measure the difference between the loss value achieved by a single-step optimization attack and an iterative one, an idea that we expand on as a possible metric. In \cite{tramerAdaptive}, the authors demonstrate that defenses that rely on gradient masking are often still vulnerable to other attacks, some of which may involve only slight tweaks to the original attack formulation. 

Some characteristic behaviours of models with masked gradients have been put forward in \cite{obfuscated}, allowing us to potentially identity them. These include:

\begin{itemize}
    \item Black box attacks outperform white box attacks
    \item Unbounded ($\epsilon = \infty$) attacks do not reach $100\%$ fooling rate
    \item Increasing the distortion bound does not decrease accuracy
\end{itemize}

While these can serve as a way to confirm whether the network under consideration suffers from gradient masking, they are quite crude tools for the task. Looking for certain characteristic behaviours in a checklist manner is inherently opaque, not providing insights into what is happening to the loss surface in gradient masked networks. Furthermore, these tests are mostly binary in output, disregarding the extent to which gradient masking takes place. This leads to a further problem in establishing insightful comparisons between models and evaluating potential proposed improvements to defenses with regard to gradient masking. These methods are also very time consuming, with black box attacks in particular tending to require excessive computational resources and being very sensitive to the chosen hyperparameters. 

The rest of the paper is structured as follows: in Section 2 we define our problem and introduce the models and defenses on which we run our experiments. Section 3 motivates our proposed metrics, and the results of our experiments are shown and discussed in Sections 4 and 5 respectively.

\section{Problem Statement}
We begin by formally setting out the terminology and notation we will use for the rest of the paper. The neural networks we describe are parametrized by $\theta$ and trained on a dataset $D$ consisting of tuples of data points $(\mathbf{x}, y)$. $\mathbf{x} \in \mathbb{R}^D$ is an image and $y$ is its true class label. We denote the label predicted by the neural network as $\hat{k}(\mathbf{x})$. The output logit for class $c$ and input $x$ is $l_c(x)$. Finally, an important quantity is the cross-entropy loss of the model, denoted as $L(x, y; \theta)$. Throughout the report, when referring to the network's gradients, we mean the gradient of the loss with respect to the input $\nabla_x L(x, y; \theta)$, which we distinguish from the gradient with respect to the network parameters $\nabla_\theta L(x, y; \theta)$ used in regular training. An adversarial example for a data tuple $(\mathbf{x}, y)$, denoted as $\mathbf{x}'$, is such that $\hat{k}(\mathbf{x}') \neq \hat{k}(\mathbf{x})$ and $\lVert \mathbf{x} - \mathbf{x}' \rVert$ is small.

Following \cite{deepfool}, we define the robustness of a network as a function of the norm of the minimal perturbation $\boldsymbol{\delta}(\mathbf{x}) \in \mathbb{R}^D$ for an image $\mathbf{x}$ such that $\mathbf{x} + \boldsymbol{\delta}(\mathbf{x})$ is an adversarial example. The robustness of a network is then defined as: $$\mathbb{E}_{(\mathbf{x}, y) \sim D} \left[ \frac{\lVert \boldsymbol{\delta}(\mathbf{x})\rVert_2}{\lVert\mathbf{x}\rVert_2}\right]$$ Desirable defenses are those that increase the robustness of a network, leading to an increase in this value.

\subsection{Model Choice}

All the experiments were done using ResNet-18 and WideResNet-28 architectures on the CIFAR-10 dataset. We chose these as ResNet architectures are the most widely used for tasks involving image datasets such as CIFAR-10.

The chosen defenses can be divided into two groups. The first are simpler methods, some of which form the basis for robust models, and are listed in Table \ref{model specs}. These are the models that were used during the development of our metrics, as their simplicity provides us with results that are more easily interpretable.  

The second group of defenses are taken from recent papers \cite{Carmon2019Unlabeled, CURE, feature_scatter, gowal2021uncovering, Wang2020Improving, Wong2020Fast,Wu2020Adversarial, zhang2020adversarial, zhang2021geometryaware, sehwag2020hydra, hendrycks2019pretraining}. In order to select these, we took into account how diverse their approaches are, as well as how well they preformed on the Autoattack \cite{autoattack} and Robustbench \cite{robustbench} leaderboards in comparison with their own reported robustness. Running our metrics on a diverse set of published defenses will show if our work is applicable and useful in current research. 


Within the first group of defenses, there are adversarial training algorithms that adapt the traditional empirical risk minimization to that shown in Equation \ref{adversarial loss}, following \cite{madry2019deep}, for a set of allowed perturbations $S$. 

\begin{equation}
    \label{adversarial loss}
    \min   \mathbb{E}_{(\mathbf{x}, y)\sim D} [\max_{\boldsymbol\delta\in S} L(\mathbf{x}+\boldsymbol\delta, y; \theta)]
\end{equation}

They effectively augment the training set with adversarial examples, with the only distinguishing variable being the attack algorithm used to approximate the max in Formula \ref{adversarial loss}. These are FGSM \cite{fgsm}, Step-ll \cite{ensemble} and PGD  \cite{madry2019deep}, each with an $\epsilon$ of 8/255 and 16/255. From these attacks, Step-ll may be less well known. It works in the same way as FGSM except, instead of maximizing the loss with respect to the target class, we minimize the loss with respect to the least likely class - the class that has the smallest logit. Single step methods are known to be poor choices for adversarial training, particularly with large epsilons. As previously pointed out, the authors in \cite{ensemble} suggest that “adversarial training with single-step methods admits a degenerate global minimum, wherein the model’s loss can not be reliably approximated by a linear function”. This would be an instance of gradient masking. PGD has been suggested as the best approximator to the point of maximum loss within an L-infinity region \cite{pgdGood}, and serves as the gold standard within this first group.

There are also those defenses that attempt to regularize the loss surface during training. They include input Jacobian regularization as in \cite{jacobian} and Gradient Clipping \footnote{Model provided by Rahul Rade}, from another ongoing project. Jacobian regularization adapts regular training by adding a regularization term to the cross-entropy loss, as shown in Equation \ref{regularization}, where $\lambda$ controls the trade off between minimizing the cross entropy loss and the regularization term and $R$ is the regularization term.

\begin{equation}
    \label{regularization}
    \min   \mathbb{E}_{(\mathbf{x}, y)\sim D} [L(\mathbf{x}, y; \theta) + \lambda R]
\end{equation}

Jacobian regularization underpins many proposals for robustness in networks, even if they are somewhat more sophisticated \cite{facebookJacobian}, making this defense an interesting one to study. We try it at different values of the $\lambda$ parameter, as described in \cite{jacobian}, which controls the trade-off between minimizing the Frobenius Norm of the Jacobian ($R$ in this case) and the cross-entropy loss. As described by \cite{hofmann}, the Frobenius Norm of the Jacobian approximates an upper bound on the maximum change in the output logits (in euclidean distance) that can arise from a small change in the inputs, making it a natural objective to minimize. However, these defenses are also prime candidates for an investigation on gradient masking, as they directly minimize the model’s gradients. In gradient clipping, the only adaptation to regular training is that the update gradients with respect to the network parameters are clipped.

\begin{table}[ht]
\center
\begin{tabular}{ll}
\toprule
                                    Model Name & Training Specification \\
\midrule
                                     ResNet-18 20 epochs & Regular training, 20 epochs \\
                                     Normal ResNet-18 & Regular trianing, 100 epochs \\
                                     Normal WideResNet-28 & Regular training, 100 epochs \\
                        Step-ll $\epsilon$: 16/255 & Adversarial training \\
                        Step-ll $\epsilon$: 8/255 & Adversarial training \\
                        FGSM $\epsilon$: 16/255 & Adversarial training \\
                        FGSM $\epsilon$: 8/255 & Adversarial training \\
                        PGD $\epsilon$: 16/255 & Adversarial training, 7 PGD iterations, relative step size = 1/4 \\
                        PGD $\epsilon$: 8/255 & Adversarial trainings, 7 PGD iterations, relative step size = 1/4 \\
                        Jacobian regularization 0.1 & Regularization training, $\lambda=0.1$ \\
                        Jacobian regularization 0.5 & Regularization training, $\lambda=0.5$ \\
                        Jacobian regularization 1 & Regularization training, $\lambda=1$ \\
                        Gradient clipping & Regular training, gradients clipped\\
\bottomrule
\end{tabular}
\caption{Models and their training specification. Adversarial training refers to Equation \ref{adversarial loss}. Regularization training refers to Equation \ref{regularization}.}
\label{model specs}
\end{table}

\subsection{Gradient Masking Benchmarks}
In order to interpret the metrics we will propose, we need to understand which of these defenses exhibit gradient masking. To do this, we make use of the behaviours described in the introduction \cite{obfuscated} as a checklist, in addition to another one we propose: there should not be a large difference between reasonable single-step gradient-based attacks and other attacks, including multi-step gradient-based attacks. We defend this as another sign of gradient masking as, if this is the case, then the direction provided by the gradient at the point is not useful. (I am not super sure about this next part). In essence, if a model is truly robust, and not simply masking its gradients, switching to a stronger attack should not lead to a large drop in accuracy.

Black-box attacks are methods that treat neural networks as opaque functions, not having access to any gradient information. They are therefore not affected by gradient masking. White-box attacks, on the other hand, typically rely on the gradients of the network to find an adversarial perturbation using optimization techniques. Black-box attacks are a subset of white-box attacks, and should be outperformed by them.
Following \cite{pgdGood}, we choose SPSA as our black-box attack, which we compare against FGSM, PGD and Step-ll as our white-box attacks. SPSA is run with 256 samples and 15 iterations. PGD is run with 25 iterations and a relative step size of $1/10$. For each of these attacks we use a distortion bound $\epsilon$ of $8/255$ and $16/255$. All of these attacks should outperform SPSA for a network without masked gradients. To check the effectiveness of unbounded attacks, we use PGD but do not project back to an L-infinity ball at any point. We expect non-gradient masked networks to have an accuracy of 0\% against this attack, as every pixel in any image can be set to a value of 0.5, impeding any classification attempt. The strongest attack is Autoattack, which we take as our state-of-the-art reference. We interpret this score as the closest thing to a true measure of robustness we can get currently. The results are shown in Table \ref{TestBenchmarks}.

We also plot FGSM vs AutoAttack accuracy in Figure \ref{FGSM AA}. We can clearly identify a correlation between FGSM and AutoAttack accuracies for models which are not masking their gradients, as we would expect. The outliers in this plot are the networks we identify as masking their gradients.

From these results, by following the checklist, we identify the models that we would expect to perform poorly on our metrics. These are both of the single-step methods trained with $\epsilon = 16/255$ and the gradient clipped model. Additionally, we observe that FGSM training with an $\epsilon = 8/255$ is extremely unstable, sometimes resulting in a phenomenon known as catastrophic overfitting(CO) \cite{Wong2020Fast}, and thus relying on gradient masking as well, so we include such an instance in our study.

As for the models from published papers, we note that Adversarial Interpolation and Feature Scatter show the same signs of gradient masking mentioned above. Notably, PGD Accuracy with an unbounded distortion bound does not reach 0\% for Adversarial Interpolation, and both models show a very large difference between FGSM and AutoAttack accuracies. The other models do not seem to suffer from this phenomenon.

\begin{table}
\begin{tabular}{cllllllllll}
\toprule
                         & Model      & Clean & \multicolumn{2}{l}{FGSM} & \multicolumn{3}{l}{PGD} & \multicolumn{2}{l}{SPSA} & AA \\
                                      &&              - &         8/255 & 16/255 &        8/255 & 16/255 & $\infty$ &         8/255 & 16/255 & 8/255 \\
\midrule
                  \parbox[t]{2mm}{\multirow{3}{*}{\rotatebox[origin=c]{90}{Undef.}}} & ResNet-18 20 epochs &           81.5 &           3.0 &    2.5 &          0.0 &    0.0 &       0.0 &          12.0 &   10.0 & 0.0 \\
                     & Normal ResNet-18 &           97.5 &          43.0 &   24.5 &          0.0 &    0.0 &       0.0 &           0.0 &    0.0 & 0.0 \\
                     & Normal WideResNet-28 &           93.0 &          28.0 &   18.5 &          0.0 &    0.0 &       0.0 &           0.0 &    0.0 & 0.0 \\      
                     \midrule
                     \midrule
              \parbox[t]{2mm}{\multirow{11}{*}{\rotatebox[origin=c]{90}{Simple Defenses}}} & Step-ll $\epsilon$: 8/255 &           74.0 &          31.5 &   13.5 &         23.5 &    4.5 &       0.0 &          41.0 &   19.0 & 21.0 \\
             & Step-ll $\epsilon$: 16/255 &           70.5 &          \textbf{57.0} &   53.5 &          \textbf{3.0} &    0.5 &       0.0 &           \textbf{8.0} &    3.0 & \textbf{1.5} \\
                 & FGSM $\epsilon$: 8/255 &           67.0 &          33.5 &   17.0 &         27.5 &    7.0 &       0.0 &          40.0 &   22.0 & 26.0 \\
            & FGSM $\epsilon$: 8/255 (CO) &           78.0 &          \textbf{73.0} &   57.5 &           \textbf{0.0} &    0.0 &       0.0 &            \textbf{2.0} &    3.0 & \textbf{0.0} \\
                & FGSM $\epsilon$: 16/255 &           71.0 &           \textbf{77.0} &   88.5 &           \textbf{0.0} &    0.0 &       0.0 &            \textbf{1.0} &    1.0 & \textbf{0.0} \\
                  & PGD $\epsilon$: 8/255 &           66.0 &          35.0 &   17.0 &         29.5 &    8.0 &       0.0 &          39.0 &   25.0 & 27.5 \\
                 & PGD $\epsilon$: 16/255 &           53.5 &          39.0 &   27.0 &         38.0 &   23.0 &       0.0 &          36.0 &   30.0 & 33.0 \\
 & Jacobian Regularization $\lambda$: 0.1 &           77.5 &          12.0 &    2.5 &          4.5 &    0.0 &       0.0 &          18.0 &    5.0 & 4.0 \\
 & Jacobian Regularization $\lambda$: 0.5 &           71.0 &          15.0 &    1.5 &          9.0 &    0.5 &       0.0 &          19.0 &    6.0 & 6.0\\
   & Jacobian Regularization $\lambda$: 1 &           63.5 &          18.5 &    3.5 &         13.0 &    1.0 &       0.0 &          22.0 &   11.0 & 10.0 \\
                            & Gradient Clipping &           93.0 &           \textbf{54.0} &   39.0 &          \textbf{11.5} &    4.0 &       0.0 &           \textbf{41.0} &   29.0 & \textbf{0.0} \\
                            \midrule
                                 \parbox[t]{2mm}{\multirow{11}{*}{\rotatebox[origin=c]{90}{Published Defenses}}} & CURE \cite{CURE} &           77.5 &          40.5 &   15.5 &         35.5 &    8.0 &       0.0 &          48.0 &   27.0 & 33.5 \\      
      & Gowal2020Uncovering \cite{gowal2021uncovering} &           88.0 &          68.0 &   51.5 &         65.0 &   37.0 &       0.0 &          70.0 &   55.0 & 61.5 \\
              & Wu2020Adversarial \cite{Wu2020Adversarial} &           88.0 &          64.0 &   47.5 &         60.5 &   34.5 &       0.0 &          63.0 &   51.0 & 57.0 \\
                  & Carmon2019Unlabeled \cite{Carmon2019Unlabeled} &           88.0 &          66.5 &   49.0 &         59.0 &   33.0 &       0.0 &          65.0 &   53.0 & 57.0 \\
                    & Wang2020Improving \cite{Wang2020Improving} &           87.5 &          67.0 &   49.5 &         62.0 &   34.0 &       0.0 &          67.0 &   55.0 & 56.5 \\
                    & Zhang2020Geometry \cite{zhang2021geometryaware} &           90.0 &          66.5 &   44.0 &         66.5 &   38.5 &       0.0 &          70.0 &   55.0 & 58.5 \\
                         & Wong2020Fast \cite{Wong2020Fast} &           83.5 &          55.0 &   31.0 &         49.5 &   15.5 &       0.0 &          60.0 &   36.0 & 42.5 \\
                   & Hendrycks2019Using \cite{hendrycks2019pretraining} &           84.5 &          61.0 &   43.5 &         56.0 &   25.5 &       0.0 &          66.0 &   50.0 & 55.0 \\
                      & Sehwag2020Hydra \cite{sehwag2020hydra} &           88.5 &          64.5 &   48.5 &         58.5 &   31.0 &       0.0 &          67.0 &   47.0 & 54.5 \\
			& Adversarial Interpolation \cite{zhang2020adversarial}&           91.5 &           \textbf{76.0} &   74.5 &         73.5 &   70.0 &       \textbf{4.0} &          73.0 &   73.0 &  \textbf{39.0} \\
			& Feature Scatter \cite{feature_scatter} &           87.0 &           \textbf{71.5} &   71.0 &         67.5 &   61.5 &       0.0 &          67.0 &   66.0 &  \textbf{39.0} \\

\bottomrule
\end{tabular}
\caption{Results for the benchmarks on testing data, shown as \% accuracy. Quantities in bold are those whose difference suggests the network in question is masking its gradients.  The midrule both groups of models discussed above.}
\label{TestBenchmarks}
\end{table}

{
\begin{figure}[t]
\centering
    \includegraphics[width=0.7\linewidth]{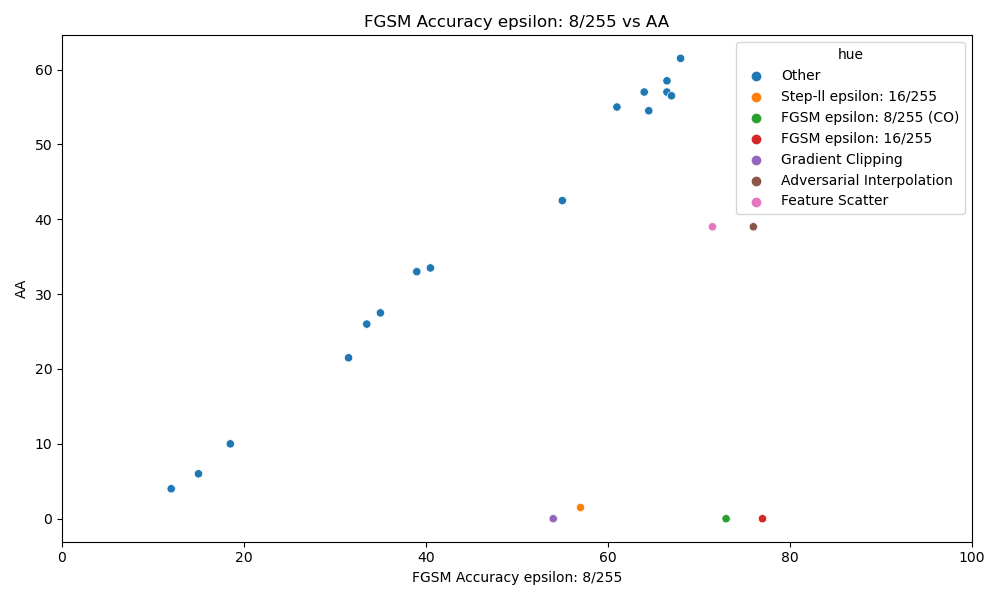}
    \caption{FGSM vs AutoAttack Accuracy.}
    \label{FGSM AA}
\end{figure}
}

These results prompted a further investigation through visualization, where we plot the values of the loss as the direction of the gradient is followed from a data point. This is done for several data points, for which the mean and standard deviation are shown in Table \ref{plotAlongGrad}, allowing us to inspect a cross-section of the loss surface. This cross-section gives us an intuition as to what is happening in gradient masked networks which will be useful in motivating our metrics. For an undefended model, as we move in the direction of the gradient, the loss quickly increases, and remains high - a gradient based attack that follows this direction will surely find an adversarial example. For a defended network with no masked gradients, following this same direction yields a much smaller increase in the loss - the fact that the decision boundary has been pushed further out from the point demonstrates that robustness has been gained. For a network with masked gradients, we see a sharp increase in the loss near the data points, followed by a very rapid decrease.  This sharp increase prompts gradient based methods to follow this direction, which ends up proving to be a very poor one to increase the loss in the long term. Thus, single-step methods, which must follow this initial direction, are left helpless. Despite thwarting single-step attacks, the SPSA results show that these models fail to truly become more robust.

These observations suggest that networks with masked gradients have abnormal loss surfaces. Measuring these abnormalities may be a way to identify the extent of gradient masking. We explore different ways to make this measurement in the next section.

\begin{table}[ht]
\centering
\begin{tabular}{ccc}
  \includegraphics[width=0.3\linewidth]{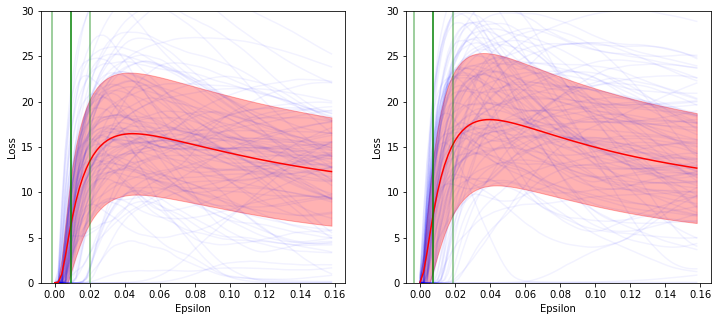} &   \includegraphics[width=0.3\linewidth]{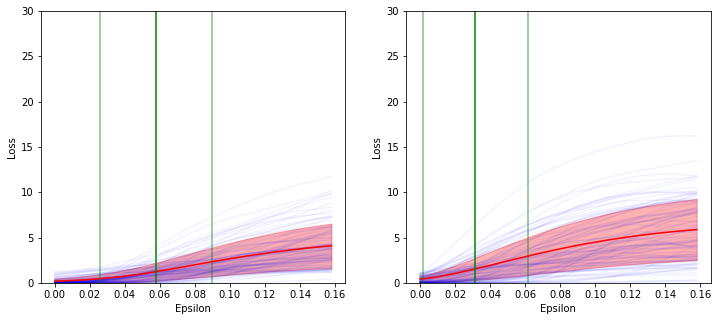} &
  \includegraphics[width=0.3\linewidth]{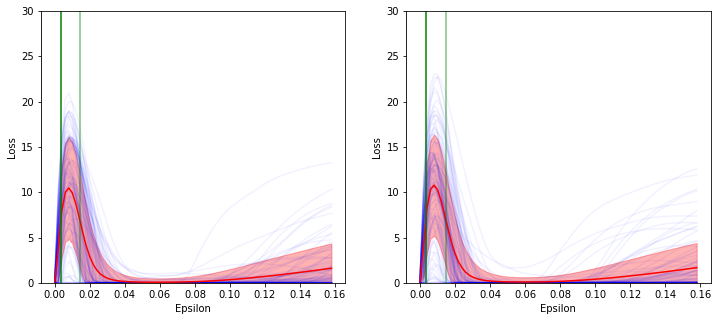}
  \\
(a) Undefended & (b) FGSM $\epsilon = 8/255$ & (c) FGSM $\epsilon = 16/255$\\[6pt]
\end{tabular}
\caption{Value of loss along the gradient direction at varying values of $\epsilon$. The mean value for each $\epsilon$ is shown in red, bounded by one standard deviation. Samples from random data points are shown in pale blue. On the left are the results for the training set, and on the right for the testing set. The solid green line shows the mean value for of the location of the decision boundary, with pale green lines showing one standard deviation in each direction}
\label{plotAlongGrad}
\end{table}

\section{Metrics}
We now motivate and formally define a series of metrics that attempt to quantify some of the characteristics pointed out in the previous section as a way to measure gradient masking.

\subsection{Gradient Norm}
The first metric we try is the mean norm of the gradients of the loss function with respect to the input. The norm of the gradient is a natural way to capture the smoothness of the loss surface at a point in the direction of the gradient. This is a crucial factor for gradient-based attacks to succeed, as they ultimately rely on linearizing the loss function. The lack of smoothness seen in Table \ref{plotAlongGrad} (c), (d), (k) should manifest itself in this measure.

Formally, we compute: $$\mathbb{E}_{(\mathbf{x}, y)\sim D} \left[ \lVert \nabla_x L(x, y) \rVert_2 \right]$$

One may think that, given a gradient with a large norm, by following its direction, and thus presumably increasing the loss, we would easily find an adversarial example. However, it is important to recall that this increase in the norm is caused by a defense which intended to make the model stronger against adversarial attacks. If the defense indeed works against at least single step gradient based attacks, we can reason that, in the majority of cases, these attacks will not find an adversarial example. We thus expect that a larger norm of gradients indicates a loss surface which is less linear, with less informative gradients.

\subsection{FGSM-PGD Cosine Similarity}
The next natural direction to explore is that of the reliability of the direction of gradients. By reliability we mean how aligned the gradient at the data point is to a vector that leads to a point of maximal loss within the distortion bound. To measure this, we can take the cosine similarity between the gradient and this optimal vector. As an approximation to the optimal direction, we settle for the perturbation provided by a PGD attack with more iterations. PGD has proven to be a very good approximator to the max operator for the loss in adversarial training, (Formula \ref{adversarial loss}) and is generally accepted as the strongest attack utilizing first order gradients\cite{pgdGood}.

Formally, with $\text{FGSM}(x, y) \in \mathbb{R}^D$ and $\text{PGD}(x, y) \in \mathbb{R}^D$ denoting the perturbation vectors returned by an FGSM and PGD attack respectively on a data point $(\mathbf{x}, y)$, we compute: $$\mathbb{E}_{(\mathbf{x, y)\sim D}} \left[ \text{Cosine Similarity} (\text{FGSM}(x, y), \text{PGD}(x, y)) \right]$$

We use a PGD attack with 10 iterations and a relative step size of $1/4$. Ultimately, this value is the cosine similarity between an FGSM attack and a PGD attack, which was been briefly mentioned by \cite{ensemble}. We extend their work by clearly motivating this value as a potential way of measuring gradient masking, and validate it with more extensive empirical testing. A small value for this metric suggests that the gradient at the data point is a poor indicator of the global loss landscape. Following its direction to increase the loss is only effective very locally, leading to the failure of single-step attacks, while high loss regions may still exist. 

\subsection{Robustness Information}
The next metric we experiment with assesses how informative gradients are at assessing the robustness at a data point. Recall from the previous section that our measure for robustness relates to the distance from a point to the nearest decision boundary. Intuitively, we want to know how accurately gradients can lead us to the nearest point on a decision boundary. To do this, we rely on the DeepFool\cite{deepfool} attack. DeepFool is a gradient based method that attempts to find the smallest perturbation that leads to an adversarial example - in other words, it approximately finds the nearest point from the data on the decision boundary. 

Although it is an approximation, in practice, this method tends to find small perturbations which are believed to be good approximations to the minimal ones. Assuming DeepFool returns the nearest point at a boundary, the normal to the decision boundary provides the optimal direction to take from the data point to reach this adversarial example. 

Let $x'$ denote the point returned by DeepFool. Additionally, let $f(z) = \nabla_{z} \left( l_{\hat{k}(x')}(z) - l_{\hat{k}(x)}(z) \right)$, where $z \in \mathbb{R}^D$. We calculate: $$\mathbb{E}_{(\mathbf{x}, y)\sim D} \left[ \text{Cosine Similarity} (f(x'), f(x)) \right]$$

Since, by definition, points $z$ at the decision boundary form a level set of $l_{\hat{k}(x')}(z) - l_{\hat{k}(x)}(z)$, $f(x')$ will be orthogonal to the boundary and point towards the adversarial class region. Thus, this measure computes how closely gradients align themselves with this "optimal" shortest path to the boundary.

\subsection{Linearization Error}
In \cite{ensemble}, the approximation-ratio between a single-step attack, Step-ll, and the true maximum loss within a region, lower bounded by PGD, is shown as an argument that adversarial training with Step-ll admits a degenerate global minimum. In the paper, they conclude that a non-linearizable loss function is an instance of gradient masking. We thus use the linearization error of the network as an explicit measure of this phenomenon.

The function $l_c$ is a multi-variable scalar valued function, and as such, we can use a first-order Taylor expansion to linearly approximate its value at any perturbation from the data point.

Let $\hat{l}_{\hat{k}(\mathbf{x})}(\mathbf{x}')$ represent the linear approximation for $l$ obtained with a taylor expansion for $\mathbf{x}' = \mathbf{x} + \boldsymbol{\delta}$ around $\mathbf{x}$. Our metric follows as: $$\mathbb{E}_{(\mathbf{x}, y)\sim D} \left[ \frac{\lvert \hat{l}_{\hat{k}(\mathbf{x})}(\mathbf{x}') - l_{\hat{k}(\mathbf{x})}(\mathbf{x}') \rvert}{\lvert l_{\hat{k}(\mathbf{x})}(\mathbf{x}') \rvert} \right]$$

Beyond proposing this as a metric for gradient masking and empirically testing it, our work extends \cite{ensemble} by directly measuring the "linearizability" of the network. Seeing as the loss is a function of the logits, this method avoids the reliance on a specific attack as well as the need for an approximation of the true maximum loss. An increase in the linearization error when a defense is applied indicates that the function learned is more irregular around the data points. While this may thwart single step attacks, the model is not necessarily more robust.

\subsection{PGD Collinearity}
When gradient masking takes place, there is a sharp curvature near the data points that causes the gradient direction to be misleading. This is clearly illustrated in Table \ref{plotAlongGrad}. Indeed, in this Figure, we see a steep decrease in the value of the loss still quite close to the data point in (c) and (d). This steep decrease, together with the fact that iterative methods such as PGD are still able to find adversarial examples using gradients, suggests that there should be a better direction, at that point, in order to increase the loss. One way to measure gradient masking could then be precisely how much iterative attacks must correct their trajectories on subsequent iterations in order to maximize the loss. 

We take this measurement by calculating the mean cosine similarity between the perturbations of consecutive iterations of PGD. If a model’s gradients are informative, we would expect the model to not deviate much from the initial gradient step, leading to a high score in this metric. More formally, let the perturbation vector returned at the i\textsuperscript{th} iteration of PGD on data point $\mathbf{x}$ with label $y$ be denoted as $p^{(i)}$. Given a PGD attack with $I$ iterations on a dataset $D$, we compute: $$ \mathbb{E}_{(\mathbf{x}, y)\sim D} \left[ \frac{1}{I-1} \sum_{i=1}^{I-1} \text{Cosine Similarity} \left(p^{(i)}, p^{(i + 1)} \right) \right]$$

\section{Results}

\begin{figure}[!htb]
  \centering
  \includegraphics[width=.7\linewidth]{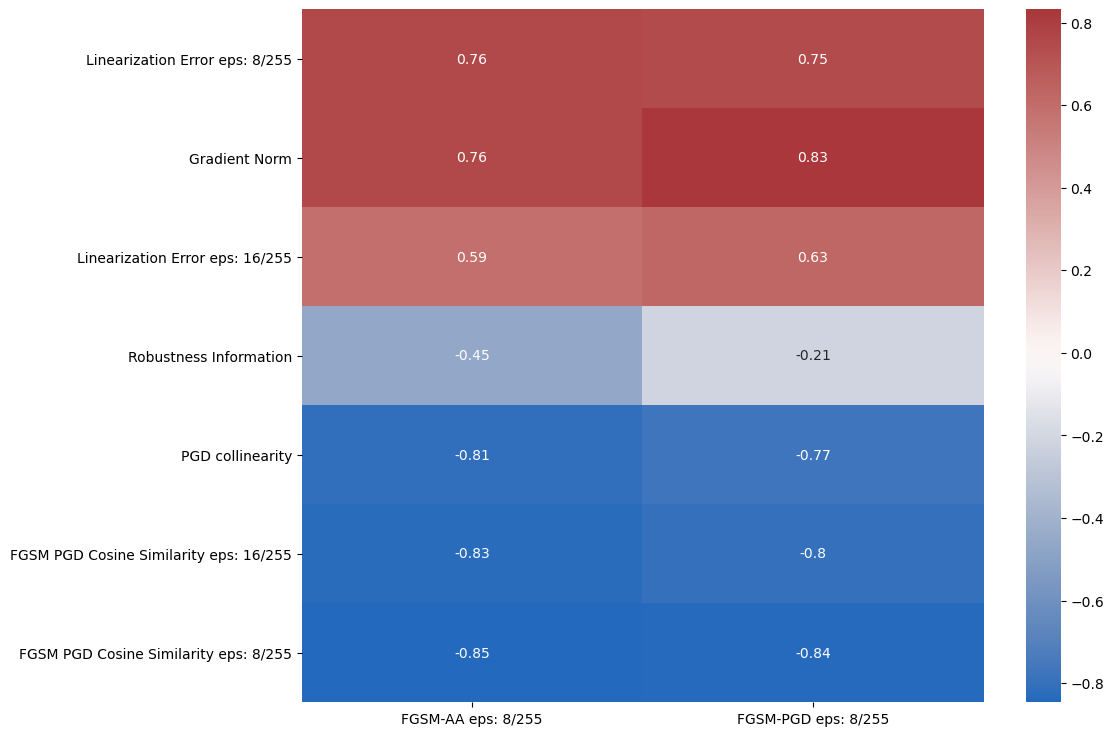}
  \caption{Correlation heatmap between metrics and quantities indicative of gradient masking. Shown for test data.}
\label{TestResultsCorr}
\end{figure}

The results can be seen in Table \ref{TestMetrics}. Large per-metric plots are shown in the appendix for better visualization. To reiterate, we are interested not only in a metrics' ability to discriminate between models that were shown to mask their gradients earlier and those that were not, but also in its capacity to illustrate the extent to which models are masking their gradients. In order to further illustrate this, we define two quantities - the difference between FGSM and AA accuracy as well the difference between FGSM and PGD accuracy. To motivate these quantities, we return to the idea we introduced in Section 2 - we should be suspicious of models that show a large difference in accuracy between reasonable single-step attacks and other, stronger attacks. We show the correlation between our metrics and these two quantities in Figure \ref{TestResultsCorr}.

\begin{table}[t]
\begin{tabular}{cllllllll}
\toprule
                         & Model       & \multicolumn{1}{C{2cm}}{Gradient Norm} & \multicolumn{2}{C{2cm}}{FGSM PGD Cosine Similarity} & \multicolumn{2}{C{2cm}}{Linearization Error} & \multicolumn{1}{C{2cm}}{PGD Collinearity} & \multicolumn{1}{C{2cm}}{Gradient Information} \\
                                      &&             - &                      8/255 &  16/255 &               8/255 &    16/255 &                - &                    - \\
\midrule
                  \parbox[t]{2mm}{\multirow{3}{*}{\rotatebox[origin=c]{90}{Undef.}}} & ResNet-18 20 epochs &        0.2544 &                     0.2905 &  0.1750 &              5.3513 &   14.0136 &           0.1694 &               0.7821 \\
                    & Normal ResNet-18 &        0.0680 &                     0.0641 &  0.0297 &              3.8535 &   30.1789 &           0.0382 &               0.3359 \\
                    &  Normal WideResNet-28 &        0.2058 &                     0.0725 &  0.0369 &              4.0258 &   60.7705 &           0.0457 &               0.4470 \\
                      \midrule
                      \midrule
             \parbox[t]{2mm}{\multirow{11}{*}{\rotatebox[origin=c]{90}{Simple Defenses}}} & Step-ll $\epsilon$: 8/255 &        0.0524 &                     0.6404 &  0.4973 &              1.0557 &    1.4191 &           0.5056 &               0.8324 \\
            & \textbf{Step-ll $\bm{\epsilon}$: 16/255} &        \textbf{0.2392} &                     \textbf{0.2214} &  \textbf{0.1255} &              \textbf{2.6540} &    \textbf{4.2211} &           \textbf{0.1959} &               \textbf{0.5433} \\
                & FGSM $\epsilon$: 8/255 &        0.0236 &                     0.6958 &  0.5505 &              0.2964 &    0.7265 &           0.5583 &               0.8390 \\
           & \textbf{FGSM $\bm\epsilon$: 8/255 (CO)} &        \textbf{0.3972} &                     \textbf{0.0669} &  \textbf{0.0466} &              \textbf{3.3727} &  \textbf{148.5161} &           \textbf{0.0642} &               \textbf{0.5629} \\
               & \textbf{FGSM $\bm\epsilon$: 16/255} &        \textbf{0.7424} &                     \textbf{0.0658} &  \textbf{0.0274} &             \textbf{11.2783} &   \textbf{12.3902} &           \textbf{0.0776} &               \textbf{0.7944} \\
                &  PGD $\epsilon$: 8/255 &        0.0204 &                     0.7190 &  0.5761 &              0.4278 &    0.5711 &           0.5858 &               0.8512 \\
                & PGD $\epsilon$: 16/255 &        0.0083 &                     0.7907 &  0.6794 &              0.2232 &    0.7346 &           0.6797 &               0.8078 \\
 & Jacobian Regularization $\lambda$: 0.1 &        0.0435 &                     0.5827 &  0.4097 &              1.1094 &    4.1042 &           0.4325 &               0.8712 \\
 & Jacobian Regularization $\lambda$: 0.5 &        0.0324 &                     0.6447 &  0.4603 &              0.5211 &    4.1131 &           0.5048 &               0.9088 \\
  & Jacobian Regularization $\lambda$: 1 &        0.0246 &                     0.6739 &  0.4985 &              0.4942 &    2.4398 &           0.5695 &               0.9074 \\
                           & \textbf{Gradient Clipping} &        \textbf{0.1014} &                     \textbf{0.1207} &  \textbf{0.0504} &              \textbf{4.6295} &   \textbf{18.4991} &           \textbf{0.0572} &               \textbf{0.1354} \\
                            \midrule
                                \parbox[t]{2mm}{\multirow{11}{*}{\rotatebox[origin=c]{90}{Published Defenses}}} & CURE \cite{CURE} &        0.0115 &                     0.7713 &  0.6288 &              0.3390 &    0.4483 &           0.6674 &               0.9045 \\

     & Gowal2020Uncovering \cite{gowal2021uncovering} &        0.0085 &                     0.7515 &  0.5623 &              0.0906 &    0.2208 &           0.6496 &               0.5349 \\
             & Wu2020Adversarial \cite{Wu2020Adversarial} &        0.0094 &                     0.7168 &  0.5352 &              0.0892 &    0.2388 &           0.6081 &               0.5969 \\
                 & Carmon2019Unlabeled \cite{Carmon2019Unlabeled} &        0.0104 &                     0.6755 &  0.4837 &              0.0924 &    0.2386 &           0.5578 &               0.5052 \\
                   & Wang2020Improving \cite{Wang2020Improving} &        0.0102 &                     0.6514 &  0.4613 &              0.1340 &    0.3150 &           0.5292 &               0.5530 \\
                   & Zhang2020Geometry \cite{zhang2021geometryaware} &        0.0104 &                     0.6755 &  0.4947 &              0.1258 &    0.3046 &           0.5626 &               0.6638 \\
                        & Wong2020Fast \cite{Wong2020Fast} &        0.0110 &                     0.6864 &  0.4883 &              0.3894 &    0.4619 &           0.5593 &               0.7331 \\
                  & Hendrycks2019Using \cite{hendrycks2019pretraining} &        0.0102 &                     0.7162 &  0.5285 &              0.1091 &    0.2439 &           0.5769 &               0.6360 \\
                     & Sehwag2020Hydra \cite{sehwag2020hydra}&        0.0104 &                     0.7010 &  0.5240 &              0.6919 &    1.7393 &           0.5842 &               0.5416 \\
                     & \textbf{Adversarial Interpolation} \cite{zhang2020adversarial} &        \textbf{0.0222} &                     \textbf{0.4473} &  \textbf{0.2935} &              \textbf{2.9747} &    \textbf{9.3566} &           \textbf{0.2951} &               \textbf{0.1579} \\
                     & \textbf{Feature Scatter} \cite{feature_scatter} &        \textbf{0.0207} &                     \textbf{0.4814} &  \textbf{0.2980} &              \textbf{2.0111} &    \textbf{8.0076} &           \textbf{0.3090} &               \textbf{0.2075} \\
\bottomrule
\end{tabular}
\caption{Results for metrics on testing data, shown as \%. Networks that were identified as masking their gradients previously are in shown in bold.}

\label{TestMetrics}
\end{table}

From Figure \ref{TestResultsCorr}, we see that both \textbf{FGSM-PGD cosine similarity} and \textbf{PGD collinearity} show quite high negative correlations with the quantities we are interested in. The networks that we identified as abnormal through the results in Table \ref{TestBenchmarks} are also singled out in the results tables, further consolidating both metrics as trustworthy. The ranking of the networks induced by these results also matches up with what we can observe in the figures, exhibiting the "capacity to illustrate the extent to which models are masking their gradients" that we are interested in. Both FGSM trained networks achieve the poorest results. They are followed by the Gradient Clipping network, as expected, and, finally, Step-ll $\epsilon$: $16/255$. For these two metrics, we plot the scores for different networks against the quantities described above in Table \ref{ResultsScatter}, to further illustrate the correlation. Beyond the correlation, the spectrum of severity of gradient masking, described in the introduction, is also revealed.

\begin{table}[t]
\centering
\begin{tabular}{cc}
  \includegraphics[width=0.5\linewidth]{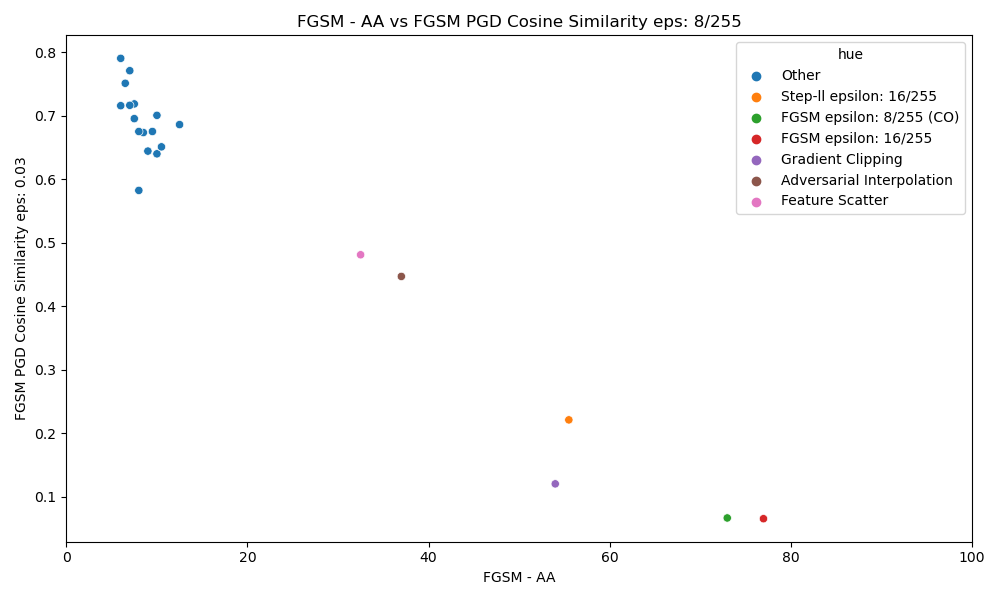} &   \includegraphics[width=0.5\linewidth]{figures/FGSM-AAvsFGSM_PGD_Cosine_Similarityeps0.03.png}
  \\
(a) FGSM PGD Cosine Similarity & (b) PGD Collinearity \\[6pt]
\end{tabular}
\caption{Scatter plots of metric score vs the difference between AA and FGSM accuracy. Shown for two selected metrics.}
\label{ResultsScatter}
\end{table}

The \textbf{Linearization Error} metric on $\epsilon = 8/255$ and \textbf{Gradient Norm} metrics show also convincing positive correlation with the quantities we defined earlier. These metrics also exhibit the same capacity to illustrate degrees of gradient masking discussed above.

The \textbf{Robustness Information} metric preforms the worst. While we do not believe the intuition behind the metric to be fundamentally flawed, there are some possible reasons for its relative failure.
One possible explanation for the poor results is its complete reliance on DeepFool. It is possible that the final result is highly dependent on how effective DeepFool is on each separate network. The second consideration is that, in models that are more robust, such as those trained using PGD, we would expect the decision boundary to be further away from the data point. Thus, it may be somewhat unfair that we are comparing the direction given at a boundary which is further away to be as reliable as that given at a boundary which is much closer to the data point. 

Overall, we consider \textbf{FGSM PGD cosine similarity} and \textbf{PGD collinearity} to be the best performing metrics. They effectively capture the extents to which networks mask their gradients, doing so with much greater computational efficiency than the initial benchmarks we compare them with. The non-linear characteristics of gradient masked networks described in \cite{ensemble, papernotPractical, obfuscated} and illustrated in figure \ref{plotAlongGrad} are also satisfactorily summarized.

\subsection{Limitations}
One limitation of these metrics is that they cannot be considered architecture agnostic - the values achieved by the same defense on a certain metric may vary according to the architecture it is employed on. Thus, for each neural network architecture being considered, a reliable benchmark from a robust model must first be established. We do not consider this a serious limitation as, realistically, there are relatively few network architectures used at the state-of-the-art across different tasks, and, once such a robust network exists, computing its score on these metrics is very fast.

(should I remove this paragraph?)
Another concern is that some of these metrics may not apply universally. Indeed, they are not based on hard theoretical evidence, but rather intuition and empirical results. For instance, while models that use gradient masking as a strategy will be less linearizable, a priori, a model that is not linearizable is not necessarily not robust. Metrics like the Linearization Error or PGD collinearity may indicate that a network's gradients are masked. However, research has shown that model robustness can be associated with models that are more "quasi-linear" around the data points \cite{CURE}. Our own results across these metrics support the idea that highly non-linear loss surfaces are in general not desirable or usual in a neural network.

\section{Conclusion} 
In this work, we study the problem of gradient masking as a quantifiable phenomenon. We set out to find metrics that are able to measure its extent across different networks. We propose a set of such metrics and test them on a set of adversarial defenses that encompass many of the trends in recently proposed defenses. These metrics are motivated intuitively and through visual explorations of the loss surface of gradient masked networks, and then extensively verified empirically. 

Supported by our empirical results, ran on published defenses, we suggest two of these metrics as useful for gradient masking: FGSM-PGD cosine similarity and PGD collinearity. We believe these metrics can be of great practical utility for the following reasons: 

\begin{itemize}
    \item They are computationally cheaper to run than expensive attacks like Autoattack
    \item they enable comparisons between models
    \item they may prevent the time-consuming task of designing a tailor-made attack for every individual defense.
\end{itemize}

We believe further work in this direction is important in order to be able to clearly identify if a network is masking its gradients or not. This would release researchers from doubting if a network is truly robust or if they simply have not been able to come up with a strong enough attack.

\clearpage
\Urlmuskip=0mu plus 1mu\relax
\printbibliography

\clearpage
\begin{appendices}
\section{Results of metrics on train data}
The following are tables for attacks and metrics on train data.

\begin{table}[!htb]
\begin{tabular}{llllllllll}
\toprule
                          Model      & Clean & \multicolumn{2}{l}{FGSM} & \multicolumn{3}{l}{PGD} & \multicolumn{2}{l}{SPSA} & AA \\
                                      &              - &         8/255 & 16/255 &        8/255 & 16/255 & $\infty$ &         8/255 & 16/255 & 8/255 \\
\midrule
\midrule
                  ResNet-18 20 epochs &          100.0 &           6.5 &    4.5 &          0.0 &    0.0 &       0.0 &           6.0 &    7.0 & 0.0 \\
                     Normal ResNet-18 &          100.0 &          44.5 &   20.0 &          0.0 &    0.0 &       0.0 &           1.0 &    0.0 & 0.0 \\
                     Normal WideResNet-28 &          100.0 &          31.0 &   22.5 &          0.0 &    0.0 &       0.0 &           0.0 &    0.0 & 0.0 \\
                     \midrule
                     \midrule
              Step-ll $\epsilon$: 8/255 &           99.5 &          70.0 &   18.0 &         44.5 &    3.5 &       0.0 &          86.0 &   32.0 & 40.5 \\
             Step-ll $\epsilon$: 16/255 &           97.5 &          98.5 &   96.5 &          2.5 &    0.5 &       0.0 &          11.0 &    2.0 & 0.5 \\
                 FGSM $\epsilon$: 8/255 &           94.5 &          77.5 &   40.5 &         69.5 &   12.0 &       0.0 &          82.0 &   62.0 & 62.5 \\
            FGSM $\epsilon$: 8/255 (CO) &           95.0 &          96.5 &   82.5 &          0.0 &    0.0 &       0.0 &           7.0 &    7.0 & 0.0 \\
                FGSM $\epsilon$: 16/255 &           79.5 &          89.0 &   98.0 &          0.0 &    0.0 &       0.0 &           1.0 &    1.0 & 0.0 \\
                  PGD $\epsilon$: 8/255 &           91.5 &          75.5 &   43.5 &         67.0 &   22.0 &       0.0 &          81.0 &   54.0 & 61.0 \\
                 PGD $\epsilon$: 16/255 &           59.0 &          46.0 &   36.5 &         45.0 &   29.0 &       0.0 &          43.0 &   34.0 & 39.5 \\
 Jacobian Regularization $\lambda$: 0.1 &           96.5 &          11.5 &    1.0 &          5.0 &    0.0 &       0.0 &          18.0 &    6.0 & 4.5 \\
 Jacobian Regularization $\lambda$: 0.5 &           83.0 &          15.5 &    2.0 &          8.0 &    0.0 &       0.0 &          14.0 &    4.0 & 6.0 \\
   Jacobian Regularization $\lambda$: 1 &           70.5 &          21.0 &    4.5 &         14.0 &    0.5 &       0.0 &          20.0 &    7.0 & 12.5 \\
                            Gradient Clipping &          100.0 &          62.5 &   50.0 &         15.5 &    3.0 &       0.0 &          38.0 &   29.0 & 0.0\\
                               \midrule
                                 CURE \cite{CURE} &           92.0 &          48.5 &   17.5 &         43.5 &   10.5 &       0.0 &          58.0 &   29.0 & 39.0 \\
                             
      Gowal2020Uncovering \cite{gowal2021uncovering} &           95.0 &          82.0 &   72.0 &         79.0 &   52.0 &       0.0 &          84.0 &   75.0 & 77.5 \\
              Wu2020Adversarial \cite{Wu2020Adversarial} &           94.0 &          83.5 &   66.0 &         80.5 &   46.5 &       0.0 &          86.0 &   73.0 & 77.0 \\
                  Carmon2019Unlabeled \cite{Carmon2019Unlabeled} &           98.0 &          85.5 &   75.0 &         82.0 &   44.5 &       0.0 &          89.0 &   78.0 & 79.0 \\
                    Wang2020Improving \cite{Wang2020Improving} &           90.5 &          76.0 &   61.5 &         72.0 &   39.0 &       0.0 &          80.0 &   67.0 & 67.0 \\
                    Zhang2020Geometry \cite{zhang2021geometryaware} &           94.5 &          83.0 &   66.0 &         80.0 &   52.0 &       0.0 &          84.0 &   77.0 & 75.5 \\
                         Wong2020Fast \cite{Wong2020Fast} &           86.0 &          59.5 &   37.5 &         53.0 &   15.0 &       0.0 &          67.0 &   40.0 & 49.5 \\
                   Hendrycks2019Using \cite{hendrycks2019pretraining} &           87.0 &          66.0 &   43.5 &         62.5 &   26.5 &       0.0 &          67.0 &   48.0 & 59.0 \\
                      Sehwag2020Hydra \cite{sehwag2020hydra} &           94.5 &          82.5 &   66.5 &         79.0 &   40.0 &       0.0 &          80.0 &   72.0 & 74.5 \\
            Adversarial Interpolation \cite{zhang2020adversarial} &          100.0 &         100.0 &  100.0 &         99.0 &   97.0 &       1.5 &         100.0 &   99.0 & 60.5 \\
                      Feature Scatter \cite{feature_scatter} &           99.5 &          98.5 &   97.0 &         97.5 &   87.5 &       0.5 &          99.0 &   99.0 & 64.0 \\
\bottomrule
\end{tabular}
\label{TrainBenchmarks}
\caption{Results for the benchmarks on training data, shown as \% accuracy. The midrule both groups of models discussed above.}
\end{table}
\begin{table}[!htb]
\begin{tabular}{lrrrrrrr}
\toprule
                                & \multicolumn{1}{C{2cm}}{Gradient Norm} & \multicolumn{2}{C{2cm}}{FGSM PGD  Cos. Sim.} & \multicolumn{2}{C{2cm}}{Linearization Error} & \multicolumn{1}{C{2cm}}{PGD Collinearity} & \multicolumn{1}{C{2cm}}{Gradient Information} \\
                                      &             - &                      8/255 &  16/255 &               8/255 &   16/255 &                - &                    - \\
\midrule
                  ResNet-18 20 epochs &        0.0001 &                     0.2894 &  0.1641 &              0.8206 &   3.6068 &           0.1734 &               0.6891 \\
                     Normal ResNet-18 &        0.0007 &                     0.0629 &  0.0279 &              1.7604 &  16.6680 &           0.0372 &               0.2084 \\
                     Normal WideResNet-28 &        0.0001 &                     0.0707 &  0.0351 &              2.2751 &  16.2693 &           0.0440 &               0.3417 \\
                     \midrule
              Step-ll $\epsilon$: 8/255 &        0.0006 &                     0.6181 &  0.4757 &              0.0973 &   0.1991 &           0.4970 &               0.7242 \\
             Step-ll $\epsilon$: 16/255 * &        0.0240 &                     0.2243 &  0.1199 &              0.4326 &   0.8501 &           0.1935 &               0.3091 \\
                 FGSM $\epsilon$: 8/255 &        0.0039 &                     0.6875 &  0.5303 &              0.0723 &   0.1547 &           0.5661 &               0.7278 \\
            FGSM $\epsilon$: 8/255 (CO) * &        0.1124 &                     0.0706 &  0.0429 &              1.0983 &   7.1962 &           0.0664 &               0.4743 \\
                FGSM $\epsilon$: 16/255 * &        0.5189 &                     0.0665 &  0.0286 &              2.3537 &  12.0186 &           0.0785 &               0.7911 \\
                  PGD $\epsilon$: 8/255 &        0.0049 &                     0.7179 &  0.5632 &              0.0708 &   0.1525 &           0.5956 &               0.7450 \\
                 PGD $\epsilon$: 16/255 &        0.0057 &                     0.7843 &  0.6734 &              0.0795 &   0.2743 &           0.6878 &               0.7739 \\
 Jacobian Regularization $\lambda$: 0.1 &        0.0134 &                     0.5981 &  0.4158 &              0.2254 &   0.6395 &           0.4389 &               0.8639 \\
 Jacobian Regularization $\lambda$: 0.5 &        0.0223 &                     0.6494 &  0.4700 &              0.1889 &   0.4389 &           0.5090 &               0.8978 \\
   Jacobian Regularization $\lambda$: 1 &        0.0234 &                     0.6787 &  0.4945 &              0.2842 &   1.0912 &           0.5700 &               0.9047 \\
                            Gradient Clipping * &        0.0001 &                     0.1093 &  0.0416 &              1.4494 &   8.7681 &           0.0565 &               0.0187 \\
                            \midrule
                                 CURE \cite{CURE} &        0.0080 &                     0.7727 &  0.6267 &              0.0794 &   0.1703 &           0.6739 &               0.8902 \\
      Gowal2020Uncovering \cite{gowal2021uncovering} &        0.0050 &                     0.7667 &  0.5714 &              0.0513 &   0.1289 &           0.6624 &               0.4558 \\
              Wu2020Adversarial \cite{Wu2020Adversarial} &        0.0053 &                     0.7201 &  0.5330 &              0.0553 &   0.1427 &           0.6168 &               0.5100 \\
                  Carmon2019Unlabeled \cite{Carmon2019Unlabeled} &        0.0045 &                     0.6932 &  0.4936 &              0.0499 &   0.1219 &           0.5780 &               0.3930 \\
                    Wang2020Improving \cite{Wang2020Improving} &        0.0073 &                     0.6594 &  0.4659 &              0.1077 &   0.2410 &           0.5386 &               0.4915 \\
                    Zhang2020Geometry \cite{zhang2021geometryaware} &        0.0065 &                     0.6826 &  0.4944 &              0.0975 &   0.2377 &           0.5715 &               0.6073 \\
                         Wong2020Fast \cite{Wong2020Fast} &        0.0082 &                     0.6851 &  0.4833 &              0.0971 &   0.2272 &           0.5589 &               0.7162 \\
                   Hendrycks2019Using \cite{hendrycks2019pretraining} &        0.0099 &                     0.7181 &  0.5304 &              0.1213 &   0.2994 &           0.5886 &               0.6379 \\
                      Sehwag2020Hydra \cite{sehwag2020hydra} &        0.0053 &                     0.7144 &  0.5233 &              0.2054 &   1.2457 &           0.5963 &               0.4668 \\
                      Adversarial Interpolation * \cite{zhang2020adversarial} &        0.0015 &                     0.5167 &  0.3549 &              0.0136 &   0.0275 &           0.3596 &               0.0063 \\
                      Feature Scatter * \cite{feature_scatter}&        0.0021 &                     0.5320 &  0.3487 &              0.0192 &   0.1383 &           0.3692 &               0.0204 \\
\bottomrule
\end{tabular}
\label{TrainMetrics}
\caption{Results for metrics on training data, shown as \%.  The midrule both groups of models discussed above.}
\end{table}

\clearpage
\section{Per Metric plots}
The following are plots with each of the metrics' scores across the different networks. They should allow for a more clear visual comparison of the metrics and models. The results are calculated with 5000 samples for each of the models. Histograms summarizing these samples are shown where they revealed distinguishing characteristics amongst the different models beyond what is summarized in the bar plots. 
\subsection{Gradient Norm}
\begin{figure}[ht]
    \centering
    \includegraphics[scale=0.35]{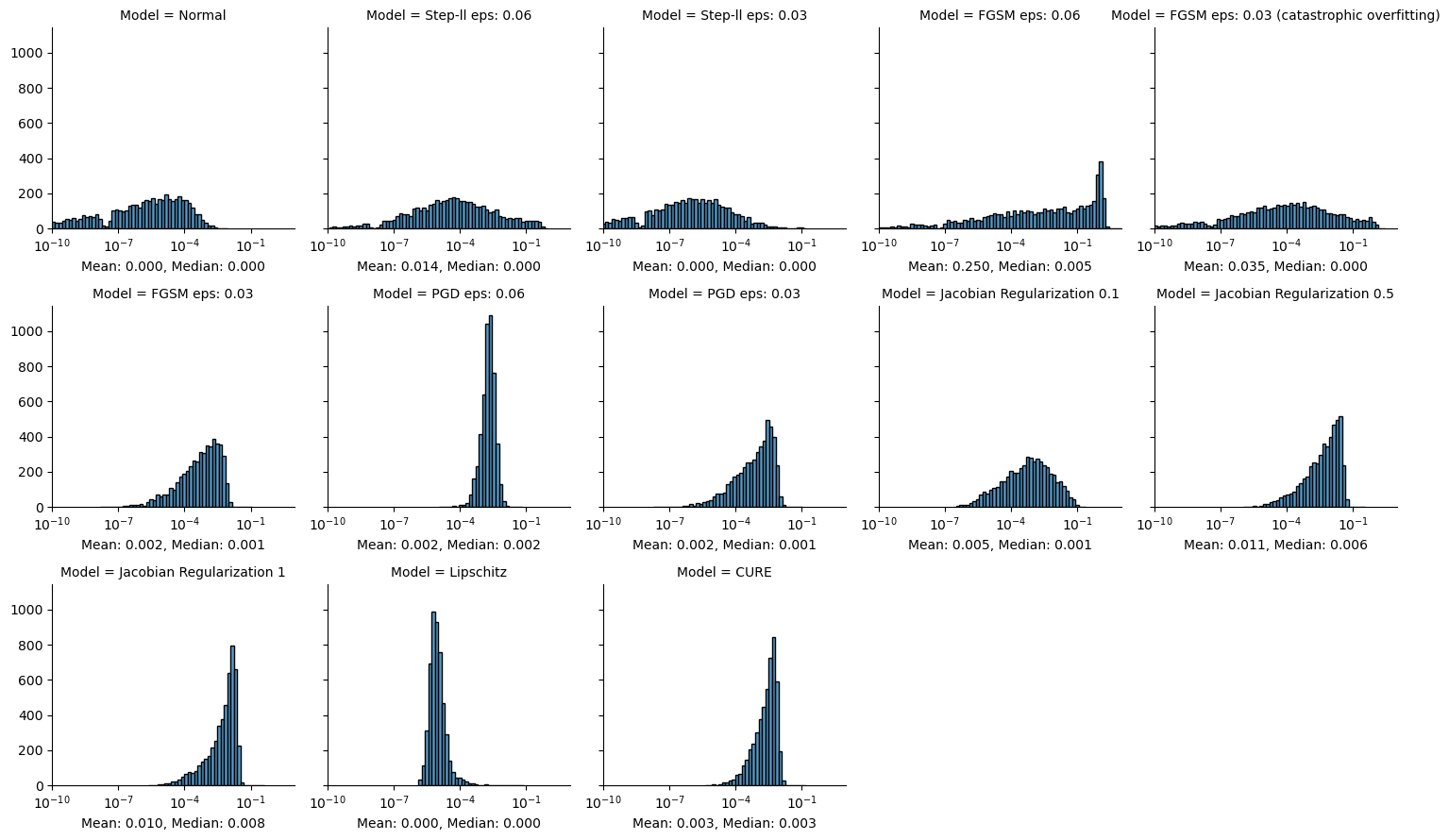}
    \caption{Gradient norm on training data}
\end{figure}

\begin{figure}[ht]
    \centering
    \includegraphics[scale=0.35]{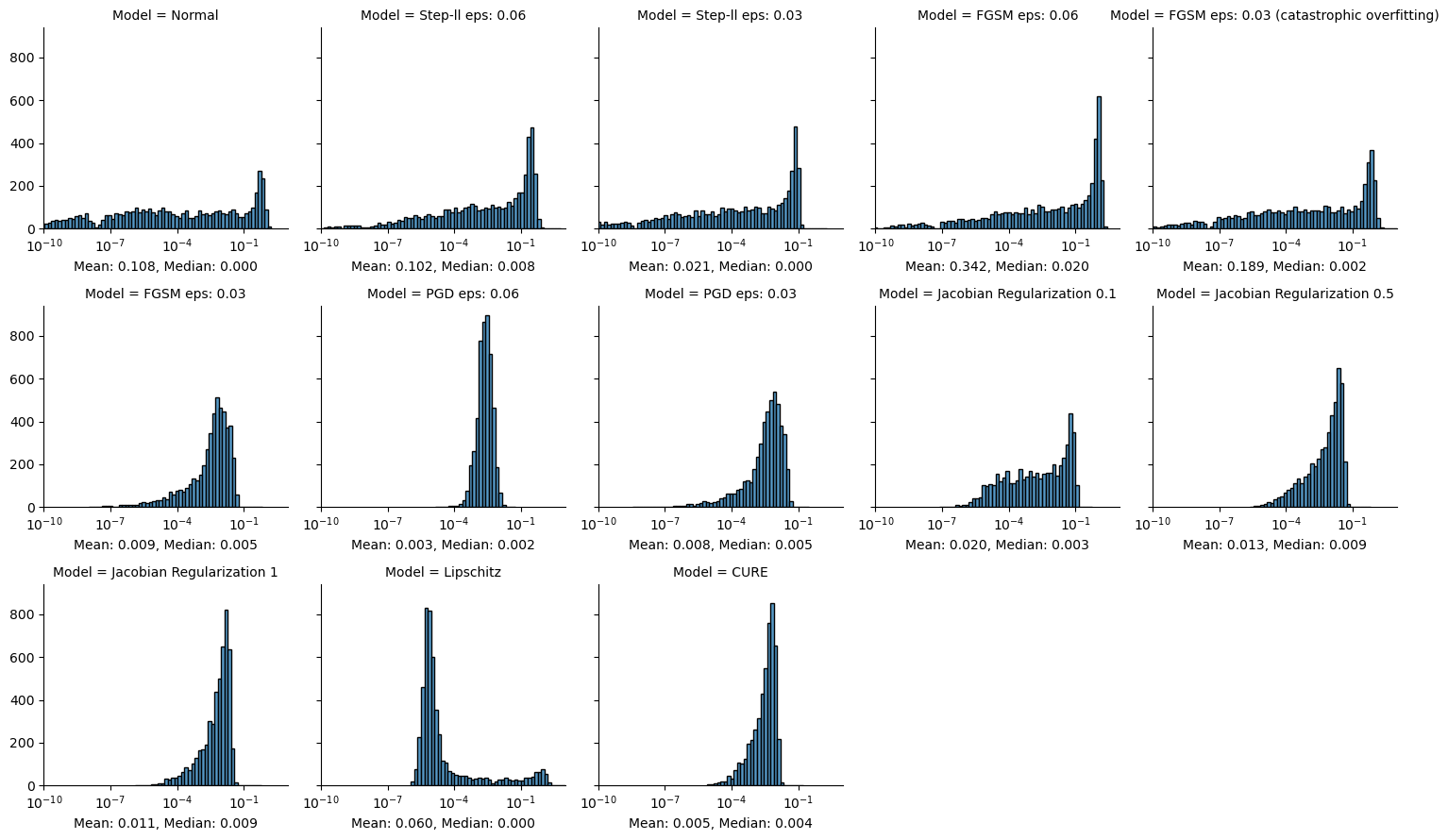}
    \caption{Gradient norm on testing data}
\end{figure}

\clearpage
\subsection{FGSM PGD Cosine Similarity}
\begin{figure}[ht]
    \centering
    \includegraphics[scale=0.35]{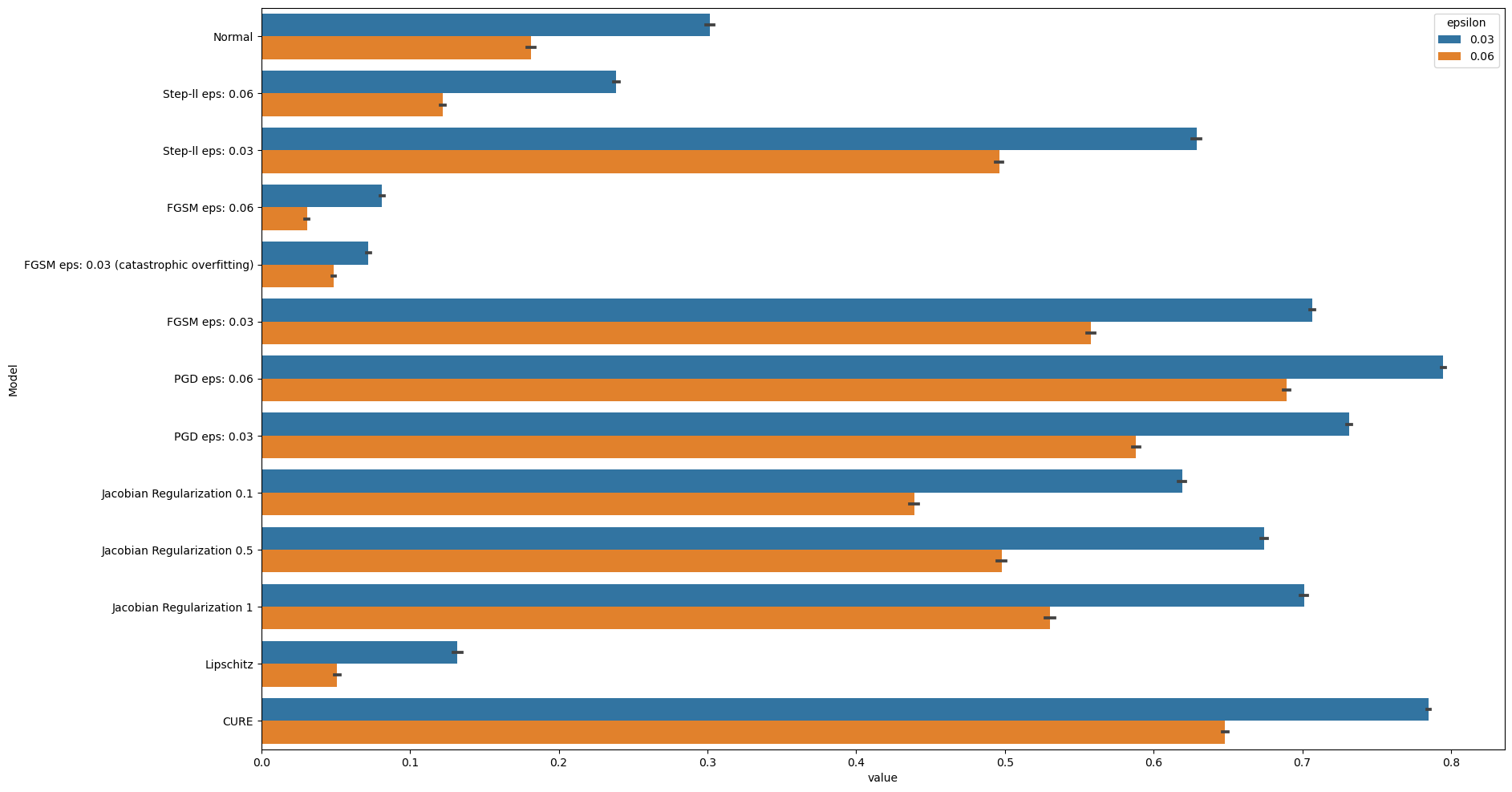}
    \caption{FGSM PGD Cosine Similarity on training data}
\end{figure}

\begin{figure}[ht]
    \centering
    \includegraphics[scale=0.35]{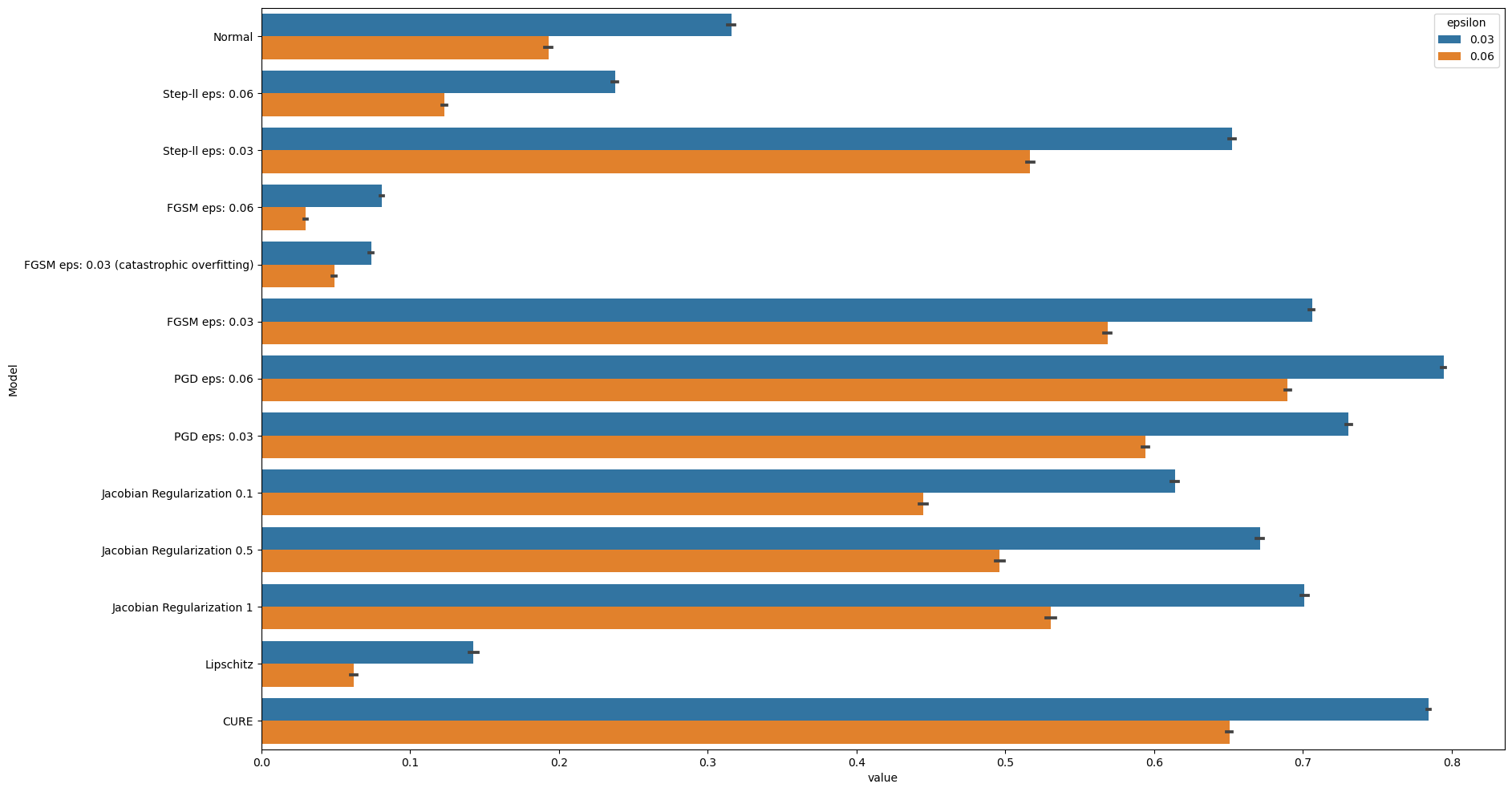}
    \caption{FGSM PGD Cosine Similarity on testing data}
\end{figure}

\clearpage
\subsection{PGD Collinearity}
\begin{figure}[ht]
    \centering
    \includegraphics[scale=0.35]{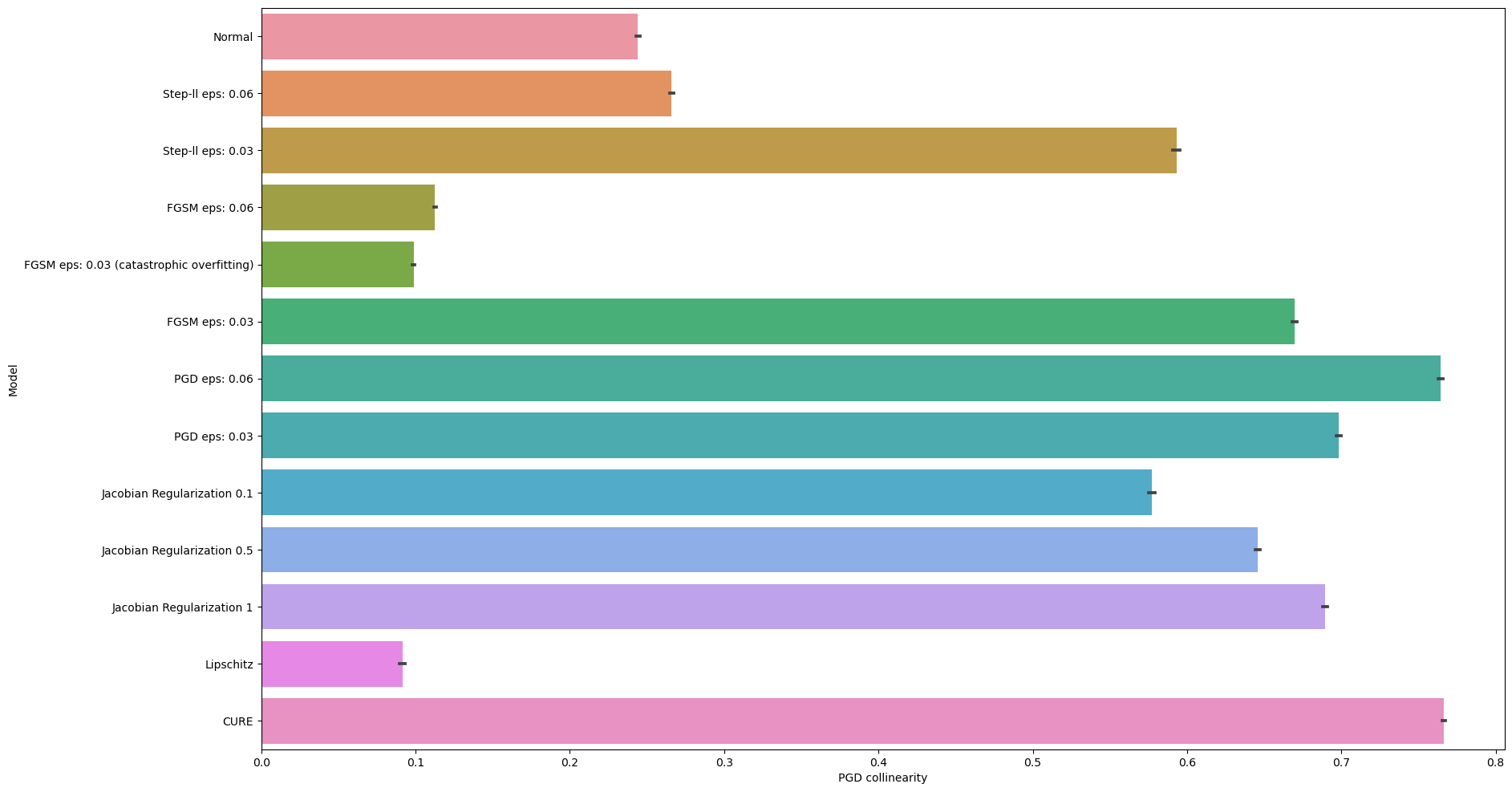}
    \caption{PGD Collinearity on training data}
\end{figure}

\begin{figure}[ht]
    \centering
    \includegraphics[scale=0.35]{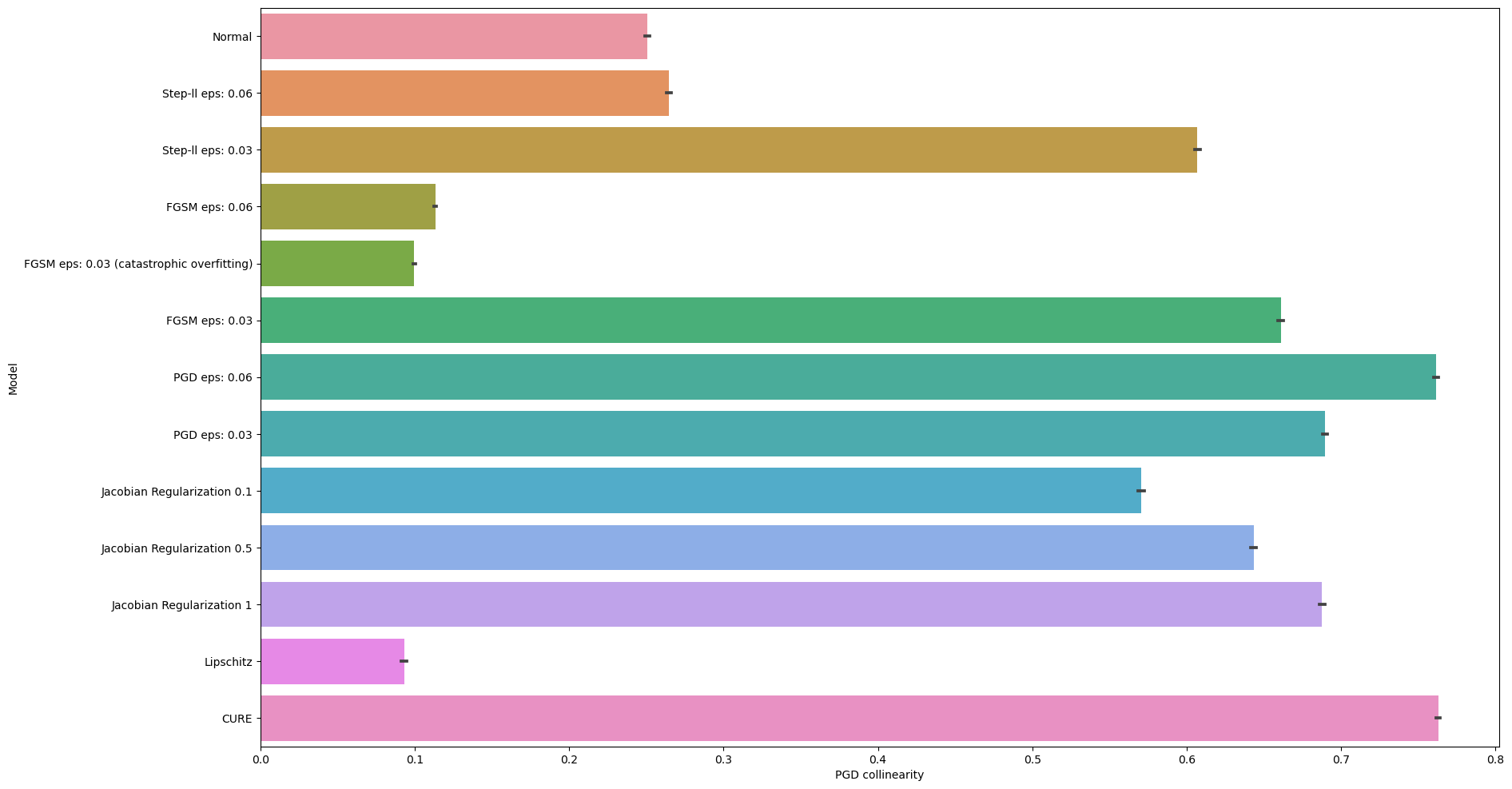}
    \caption{PGD Collinearity on testing data}
\end{figure}

\clearpage
\subsection{Robustness Information}
\begin{figure}[ht]
    \centering
    \includegraphics[scale=0.35]{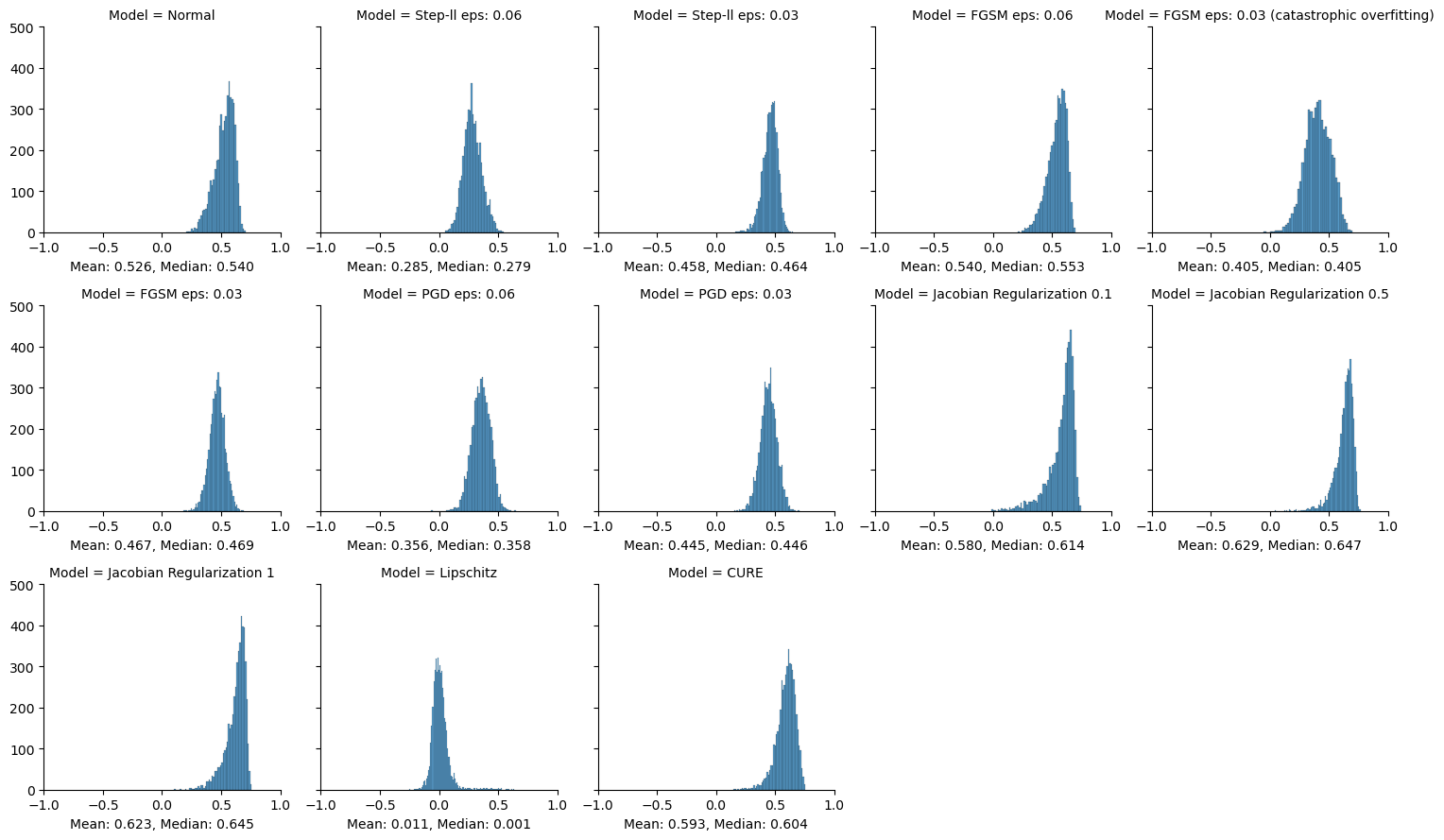}
    \caption{Robustness information on training data}
\end{figure}

\begin{figure}[ht]
    \centering
    \includegraphics[scale=0.35]{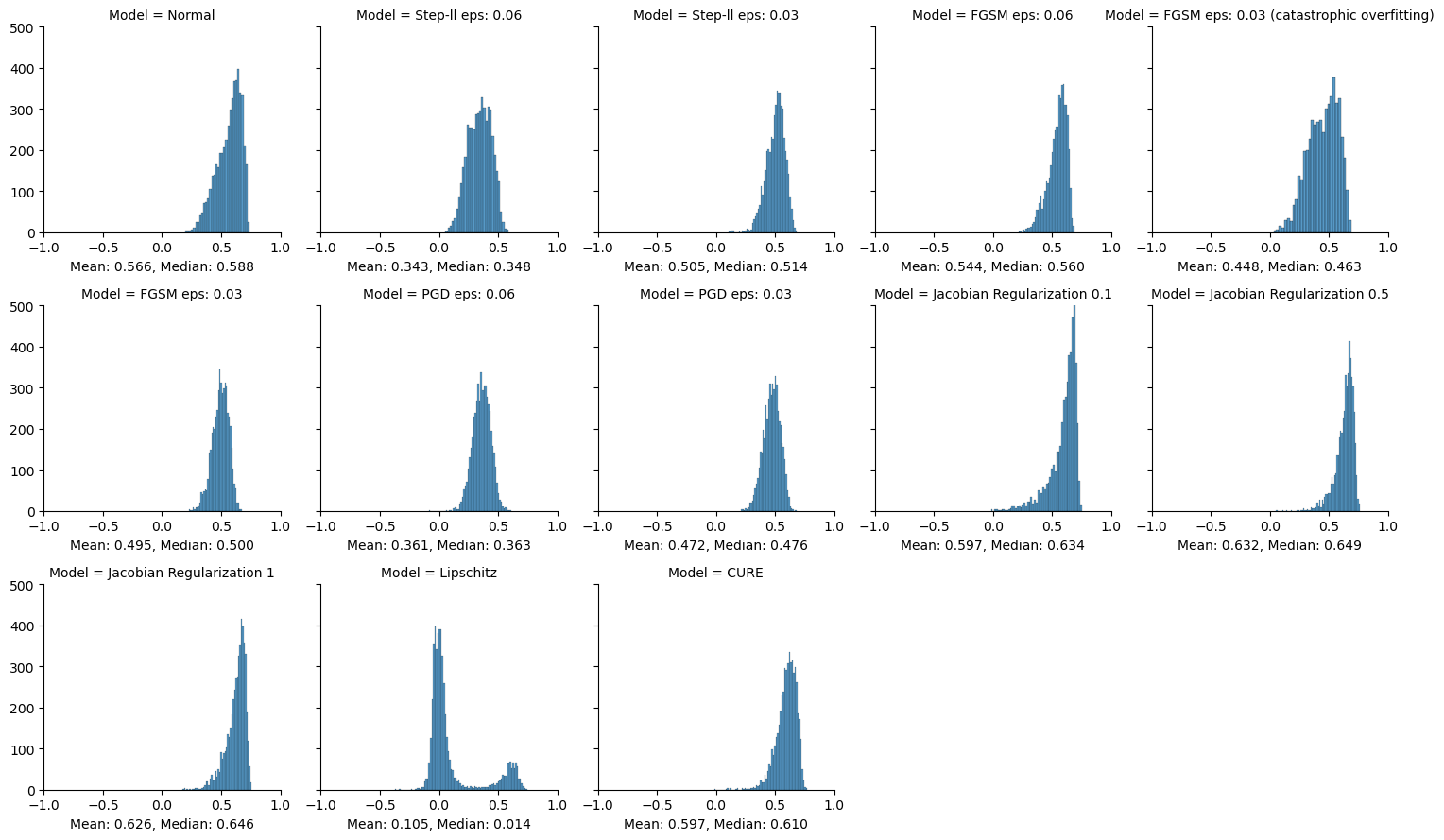}
    \caption{Robustness information on testing data}
\end{figure}

\clearpage
\subsection{Linearization Error}
\begin{figure}[ht]
    \centering
    \includegraphics[scale=0.35]{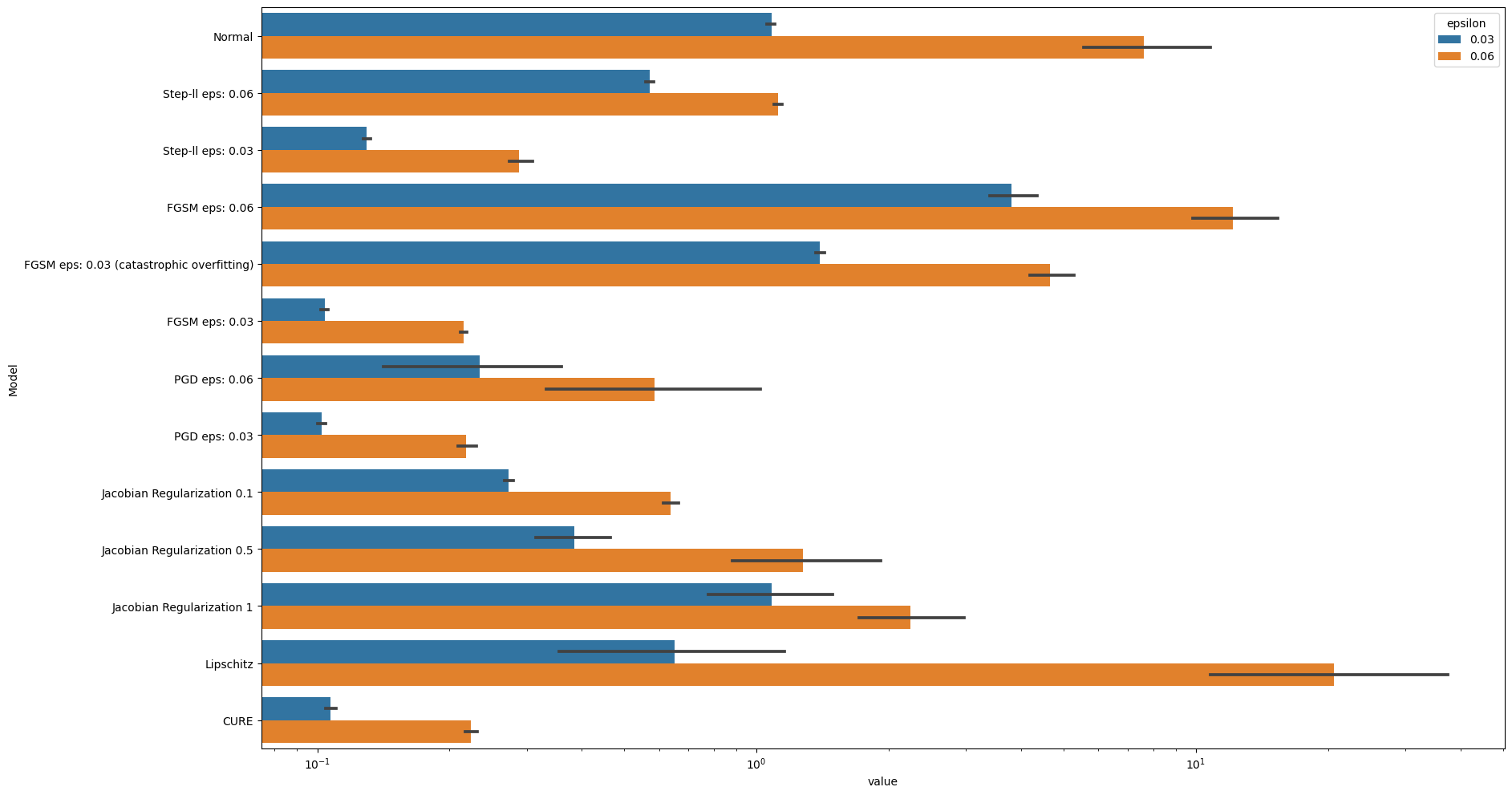}
    \caption{Linearization error on training data}
\end{figure}

\begin{figure}[ht]
    \centering
    \includegraphics[scale=0.35]{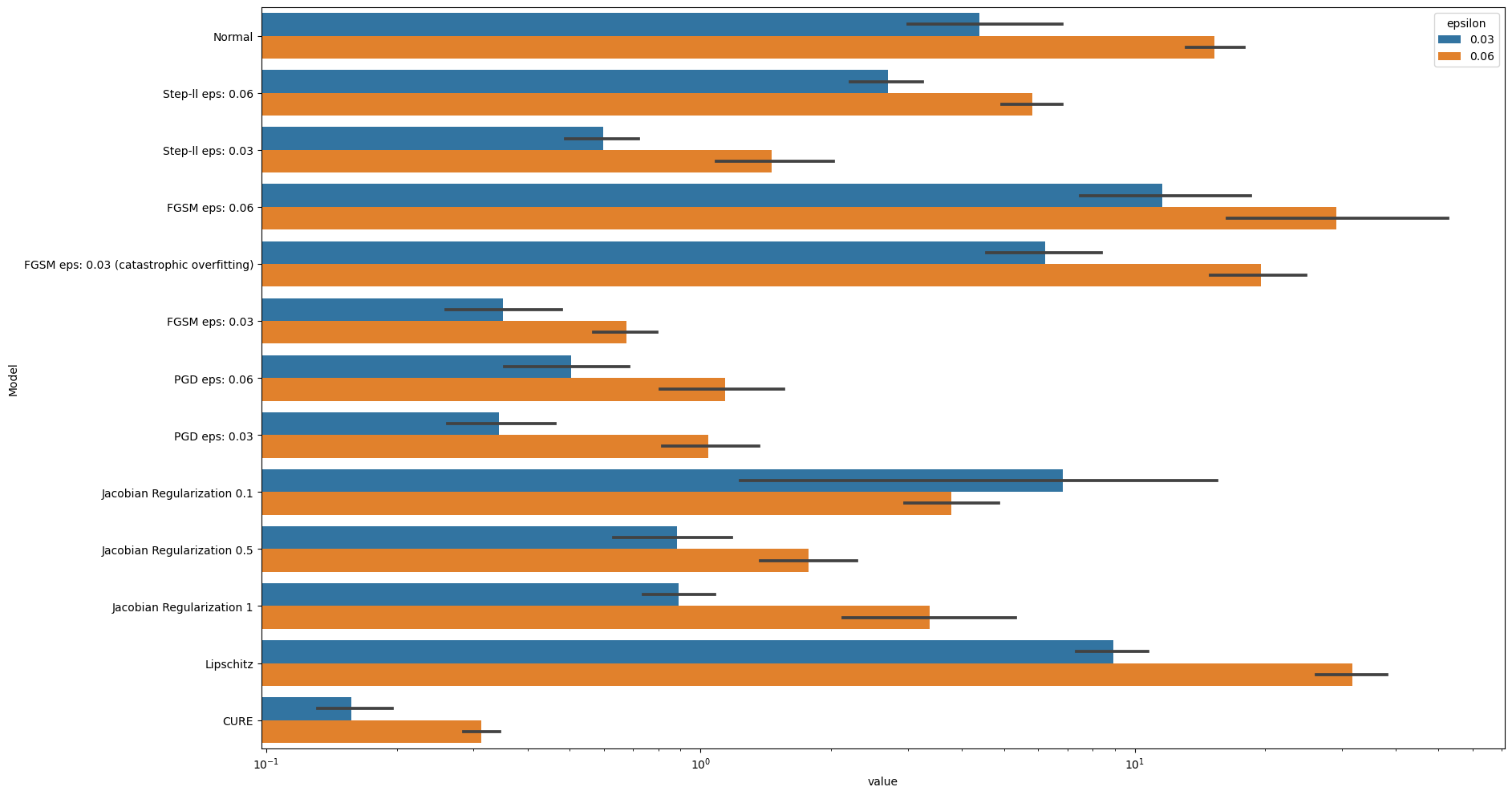}
    \caption{Linearization error on testing data}
\end{figure}
\end{appendices}
\end{document}